\documentclass[letterpaper, 10 pt, conference]{ieeeconf}  

\IEEEoverridecommandlockouts                              
\usepackage{graphicx} 
\usepackage{amsmath} 
\usepackage{amssymb}  
\usepackage{xfrac}
\usepackage{gensymb}
\usepackage{subfig}
\let\labelindent\relax
\usepackage[shortlabels]{enumitem}

\bstctlcite{IEEEexample:BSTcontrol}

\title{\LARGE \bf
Passive Quadrupedal Gait Synchronization for Extra Robotic Legs Using a Dynamically Coupled Double Rimless Wheel Model}

\author{Daniel J. Gonzalez, \IEEEmembership{Member,~IEEE}$^{1}$, and H. Harry Asada, \IEEEmembership{Member,~IEEE}$^{2}$
\thanks{$^{1}$D. J. Gonzalez is with both the Robotics Research Center in the Department of Electrical Engineering and Computer Science at the United States Military Academy, West Point, NY 10996, USA and the d'Arbeloff Laboratory for Information Systems and Technology in the Department of Mechanical Engineering, Massachusetts Institute of Technology, Cambridge, MA 02139, USA. Email: {\tt\small daniel.gonzalez@westpoint.edu} and {\tt\small dgonz@mit.edu}}
\thanks{$^{2}$H. H. Asada is with the d'Arbeloff Laboratory for Information Systems and Technology in the Department of Mechanical Engineering, Massachusetts Institute of Technology, Cambridge, MA 02139, USA. Email: 
        {\tt\small asada@mit.edu}}%
}

\begin{document}
\bstctlcite{IEEEexample:BSTcontrol}

\maketitle
\thispagestyle{empty}
\pagestyle{empty}

\begin{abstract}
The Extra Robotic Legs (XRL) system is a robotic augmentation worn by a human operator consisting of two articulated robot legs that walk with the operator and help bear a heavy backpack payload. It is desirable for the Human-XRL quadruped system to walk with the rear legs lead the front by 25\% of the gait period, minimizing the energy lost from foot impacts while maximizing balance stability. Unlike quadrupedal robots, the XRL cannot command the human's limbs to coordinate quadrupedal locomotion. Using a pair of Rimless Wheel models, it is shown that the systems coupled with a spring and damper converge to the desired 25\% phase difference. A Poincar\'e return map was generated using numerical simulation to examine the convergence properties to different coupler design parameters, and initial conditions. The Dynamically Coupled Double Rimless Wheel system was physically realized with a spring and dashpot chosen from the theoretical results, and initial experiments indicate that the desired synchronization properties may be achieved within several steps using this set of passive components alone.

\end{abstract}


\section{Introduction and Motivation}\label{Intro}
The Extra Robotic Legs (XRL) system aims to empower the industrial worker and emergency responder to enhance their ability to perform their job by alleviating the burden of heavy equipment and enabling them to execute strenuous maneuvers more easily. The XRL system was designed to allow United States Department of Energy (DOE) nuclear decommissioning workers to carry more life support equipment (such as an air tank, extra tools, and a body cooling system) to increase their ``stay time'' at the task location, and to support workers in general who must take kneeling, crouching, and other fatiguing postures near the ground \cite{Gonzalez2018}. Previous work has explored the shared control of balance while the operator squats down to the ground \cite{Gonzalez2019}. We now investigate the synchronization of the human and XRL bipedal systems with each other during steady-state locomotion.

\begin{figure}[ht]
	\centering
	\includegraphics[width=0.75\linewidth]{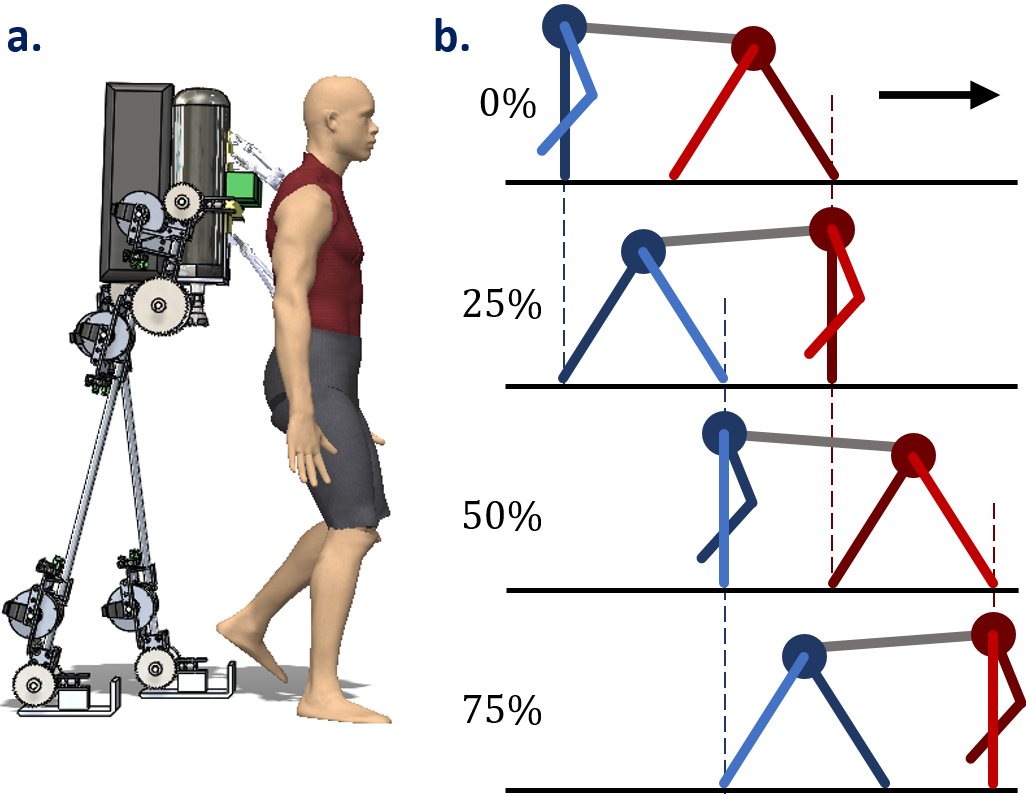}
	\caption{The Extra Robotic Legs System and desired gait cycle, with the hind legs leading the rear legs by 25\%. This quadrupedal system behaves as if it were two coupled bipeds.}
	\label{figGaitCycle}
\end{figure}

When walking together, the human-XRL system forms a type of quadrupedal system (See Fig. \ref{figGaitCycle}-a). Unlike fully biological or fully robotic quadrupeds, the human-XRL system consists of two independently controlled biped systems which are physically connected; one is a human and the other is a robot. The synchronization challenge arises from the fact that there is no centralized controller to command the entire quadruped. Our goal is to establish a natural regulator that achieves a desired gait cycle by exploiting the intrinsic dynamic synchronization properties of the human-XRL system.

This desired gait cycle comes from studying animal biomechanics. Analysis of quadrupedal animal gaits \cite{Griffin2004} shows that quadrupeds behave as if they were two coupled bipeds and tend to fall into a gait cycle where the hind limbs lead the fore limbs by about 25\% of the stride time (or $90^\circ$ out of phase) during steady-state walking, as shown in Fig. \ref{figGaitCycle}-b. This walking gait cycle in which the footfalls are timed in a 1-4-2-3 sequence (numbered clockwise starting from the front-left foot) has also been found to maximize the margin of stability of the quadruped's balance \cite{McGhee1968}. It has also been shown that gaits with more sequenced collisions per stride (such as in the four-beat walking gait) are more energy efficient than gaits which group multiple foot collisions together \cite{Ruina2005}. Using a quadrupedal model incorporating legs with mass and a wobbling mass passively connected to a rigid body, the four-beat gait can be stabilized \cite{Remy2010}. It is of interest, then, to analyze how intrinsic dynamics can lead the Human-XRL System to naturally fall into this special gait cycle. 

In this work, we aim to establish an interactive passive gait synchronization method for the human-XRL system. In Section \ref{coupledModel} we introduce and analyze a simple, yet novel, quadrupedal walking model consisting of two Rimless Wheel biped systems connected with a dynamic coupler: the Dynamically Coupled Double Rimless Wheel model. Property-agnostic coupler conditions that maximize the convergence rate are found using a numerically acquired Poincar\'e map in Section \ref{PoincareAnal}. Synchronization is demonstrated on a physically realized Coupled Rimless Wheel pair in Section \ref{Experiment}. Section \ref{Conclusion} provides a conclusion and outlines future work to be conducted. 

\section{The Dynamically Coupled Double Rimless Wheel Model}\label{coupledModel}
The canonical Rimless Wheel model \cite{McGeer1990} is a simple model that captures natural bipedal walking dynamics. The Rimless Wheel model has been extended to capture quadrupedal walking dynamics by coupling a pair of Rimless Wheels with a rigid beam and forcing a set footfall timing sequence \cite{Smith1997}. Stable walking limit cycles can be observed with this model, though the gait cycle phase is strictly enforced through kinematics. Such a system was then shown to maximize passive steady-state velocity when the phase parameter was chosen to be perfectly out of phase, or 50\% out of phase\footnote{Note that in quadrupedal locomotion we consider the system to be at our desired gait cycle when the hind legs are 25\% out of phase with the front, but in planar models there is no distinction between the left or right legs, and thus we consider successful synchronization to be achieved when the stance leg of the front and rear halves are 50\% out of phase from each other. The additional virtual pair of legs in the swing phase are not taken into account in this model.} \cite{Inoue2011}. We now introduce some novel modifications of our own. 

In order to explore the effects of dynamic coupling, we extend the Rimless Wheel model by adding a second Rimless Wheel and connecting the two with a passive coupler made up of a spring $k$ and a viscous damper $b$ in parallel.
\begin{figure}[ht]
	\centering
	\includegraphics[width=0.75\linewidth]{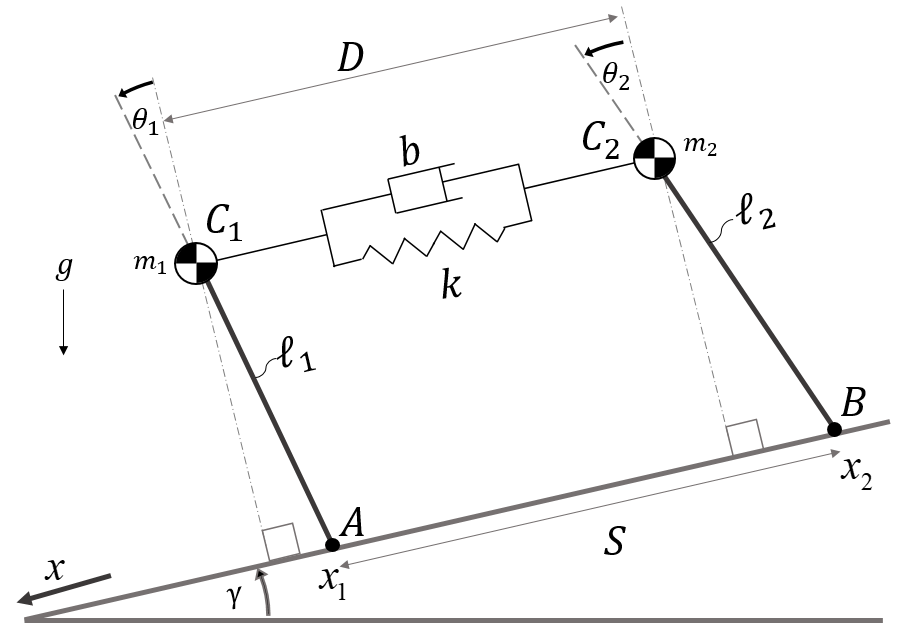}
	\caption{The Coupled Rimless Wheels Model}
	\label{coupledrimlesswheelfig}
\end{figure} 

Characteristic to our human-XRL system is that this coupler is dynamic rather than rigid. Smith and Berkmeier use a rigid connection, which kinematically couples the angles $\theta_1$ and $\theta_2$ and the angular velocities $\dot{\theta}_1$ and $\dot{\theta}_2$ \cite{Smith1997}. Once a set of initial conditions has been chosen, the rigidly coupled Rimless Wheels cannot change their relative phase. 

Our novel model, by including a passive spring and damper connecting the two wheels, allows the angles and angular velocities of each wheel to move independently, allowing for rich dynamic interactions between the two. Tuning design parameters $k$ and $b$ can affect the mutual dynamics and gait cycle. Indeed, it is through this mechanism that we aim to synchronize the gait cycle of the first and second Rimless Wheels.

The masses $m_1$ and $m_2$ of each pendulum are point masses atop massless links of length $\ell_1$ and $\ell_2$, respectively. The angle of the first Rimless Wheel about point $A$ is $\theta_1$ and the angle of the second about point $B$ is $\theta_2$, and the step angles for each are $\alpha_1$ and $\alpha_2$. The distance between the coupler endpoints is $D$ and the unstretched length of the spring is $D_0$. See Fig. \ref{coupledrimlesswheelfig}.

The force balance equations for the Rimless Wheels are
\begin{equation}
    \begin{split}
        m_1\ell_1^2\ddot{\theta}_1=&m_1g\ell_1\sin{(\theta_1+\gamma)}\\
        -&F_c\ell_1(\cos{\theta_1}\cos{\beta}-\sin{\theta_1}\sin{\beta})        
    \end{split}
\end{equation}
\begin{equation}
    \begin{split}
        m_2\ell_2^2\ddot{\theta}_2=&m_2g\ell_2\sin{(\theta_2+\gamma)}\\
        +&F_c\ell_2(\cos{\theta_2}\cos{\beta}-\sin{\theta_2}\sin{\beta})
    \end{split}
\end{equation}
where the coupler force is
\begin{equation}
F_c = k(D-D_0)+b\dot{D}
\end{equation}

\begin{figure}[ht]
	\centering
	\includegraphics[width=\linewidth]{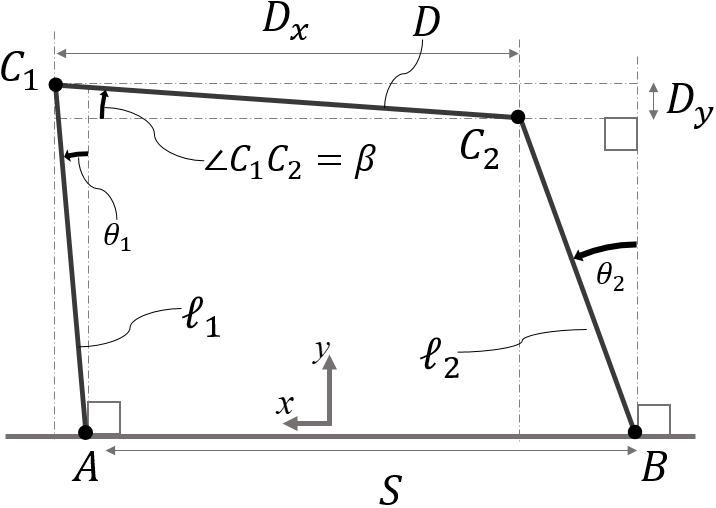}
	\caption{Geometric Relationship for $D$ and $\beta$.}
	\label{DxDyBetafig}
\end{figure} 

The coupler of length $D$ acts between the two pendula and has an orientation $\beta$ relative to the sloped ground surface. See Fig. \ref{DxDyBetafig} for details. The $x$-component of $D$ is
\begin{equation}
D_x =D\cos\beta= \ell_1 \sin{\theta_1} - \ell_2 \sin{\theta_2} + S
\end{equation}
and the $y$-component of $D$ is
\begin{equation}
D_y =D\sin\beta= \ell_1 \cos{\theta_1} - \ell_2 \cos{\theta_2}
\end{equation}
where $S$ is the distance between the stance foot locations $x_1$ and $x_2$ of each Rimless Wheel $i$
\begin{equation}
S = x_1 - x_2
\end{equation}
From these components, we find the coupler distance 
\begin{equation}
D=\sqrt{D_x^2+D_y^2}
\end{equation}
the coupler angle
\begin{equation}
        \beta
        =\sin^{-1}\left(\frac{\ell_1 \cos{\theta_1} - \ell_2 \cos{\theta_2}}{D}\right)    
\end{equation}
and the coupler distance time derivative
\begin{equation}
\dot{D}=\frac{D_x\dot{D}_x+D_y\dot{D_y}}{D}
\end{equation}
where
\begin{equation}
\dot{D}_x=\ell_1\cos{(\theta_1)}\dot{\theta}_1-\ell_2\cos{(\theta_2)}\dot{\theta}_2
\end{equation}
and
\begin{equation}
\dot{D}_y=\ell_2\sin{(\theta_2)}\dot{\theta}_2-\ell_1\sin{(\theta_1)}\dot{\theta}_1
\end{equation}
resulting in
\begin{equation}
\dot{D}=\ell_1\cos(\theta_1+\beta)\dot{\theta}_1-\ell_2\cos(\theta_2+\beta)\dot{\theta}_2
\end{equation}


The stance distance $S$ is piecewise constant
\begin{equation}
S = S_0+ n_1\sigma_1 - n_2 \sigma_2
\end{equation}
and depends on the initial step length $S_0$, the number of steps taken by each wheel $n_i$ and the step length $\sigma_i$ for each Rimless Wheel
\begin{equation}
\sigma_i = 2\ell\sin{\alpha_i}, ~i\in1,2
\end{equation}
If step lengths are the same ($\sigma_1 = \sigma_2 = \sigma$) then
\begin{equation}
S = S_0 + n\sigma
\end{equation}
where
\begin{equation}
n = n_1-n_2
\end{equation}
is the relative difference in the number of steps between the front and back halves. During normal operation, $n$ is $0$ or $1$.


The nonlinear state-determined equations during the continuous dynamics are, assuming $\ell_1=\ell_2=\ell$ and $m_1 = m_2 = m$:
\begin{equation}\label{eqSS1}
    \begin{split}
        \ddot{\theta}_1=&\frac{g}{\ell}\sin{(\theta_1+\gamma)}\\
        -&\frac{k}{m\ell}(D-D_0)\cos{(\theta_1+\beta)}\\
        -&\frac{b}{m}\left(\cos(\theta_1+\beta)\dot{\theta}_1-\cos(\theta_2+\beta)\dot{\theta}_2\right)\cos{(\theta_1+\beta)}
    \end{split}
\end{equation}
\begin{equation}\label{eqSS2}
    \begin{split}
        \ddot{\theta}_2=&\frac{g}{\ell}\sin{(\theta_2+\gamma)}\\
        +&\frac{k}{m\ell}(D-D_0)\cos{(\theta_2+\beta)}\\
        +&\frac{b}{m}\left(\cos(\theta_1+\beta)\dot{\theta}_1-\cos(\theta_2+\beta)\dot{\theta}_2\right)\cos{(\theta_2+\beta)}
    \end{split}
\end{equation}
\begin{equation}\label{eqSS3}
    \dot{D}=\ell\left(\cos(\theta_1+\beta)\dot{\theta}_1-\cos(\theta_2+\beta)\dot{\theta}_2\right)
\end{equation}
where, for brevity, we write
\begin{equation}
\beta=\sin^{-1}\left(\frac{\ell}{D}(\cos\theta_1 - \cos\theta_2)\right)
\end{equation}
and the state of the system can be fully determined with the following state vector
\begin{equation}
x = \begin{bmatrix}
\theta_1&
\dot{\theta}_1&
\theta_2&
\dot{\theta}_2&
D
\end{bmatrix}^T
\end{equation}
where each of the five energy storage elements in the system (kinetic energy of each pendulum, gravitational potential energy of each pendulum, and potential energy stored in the spring) is associated with its own state variable. 

The hybrid heel strike/toe-off dynamics are treated independently for each pendulum system at the angle limit $\alpha$ of forward lean before the swing leg impacts and becomes the new stance leg. Energy is lost by conserving only angular momentum about the new stance leg. The discrete jumps for the hybrid dynamical system occur as follows (assuming near-steady-state rolling in the positive $\theta$ direction):
\begin{enumerate}[A.]
	\item when $\theta_1 \geq \alpha$:\\ 
	The angle instantaneously changes: $\theta_{1+}=-\alpha$; \\
	The angular velocity instantaneously changes: $\dot{\theta}_{1+}=\dot{\theta}_{1-}\cos{(2\alpha)}$; \\
	The difference in steps between Rimless Wheels 1 and 2 increases: $n = 1$.
	\item when $\theta_2 \geq \alpha$:\\ 
	The angle instantaneously changes: $\theta_{2+}=-\alpha$;\\
	The angular velocity instantaneously changes: $\dot{\theta}_{2+}=\dot{\theta}_{2-}\cos{(2\alpha)}$;\\
	The difference in steps between Rimless Wheels 1 and 2 decreases: $n = 0$.
\end{enumerate}

Note that $D$ is continuous and does not ever jump discretely, but $\dot{D}$ does abruptly jump due to its proportionality to $\dot{\theta}_1$ and $\dot{\theta}_2$. 

\section{Coupler Parameters and Synchronization Rate}\label{PoincareAnal}
To generalize our numerical analysis of the properties of the Coupled Rimless Wheels System, we nondimensionalize the state equations. 
By defining a unit time
\begin{equation}
\tau=t\sqrt{\ell/g}
\end{equation}
and choosing nondimensionalized parameters, initial conditions, and state variables
\begin{equation}\label{paramNonDim}
\hat{D}=\frac{D}{\ell}, ~\hat{D_0}=\frac{D_0}{\ell}, ~\hat{k}=\frac{k}{mg}, ~\hat{b}=\frac{b}{m}\sqrt{\frac{\ell}{g}}
\end{equation}
the nondimensionalized nonlinear state equations are
\begin{equation}
    \begin{split}
        \theta_1''=&\sin{(\theta_1+\gamma)}-\hat{k}(\hat{D}-\hat{D}_0)\cos{(\theta_1+\beta)}\\
        -&\hat{b}\left( \cos{(\theta_1+\beta)}\theta_1' - \cos{(\theta_2+\beta)}\theta_2'\right)\cos{(\theta_1+\beta)}
    \end{split}
\end{equation}
\begin{equation}
    \begin{split}
        \theta_2''=&\sin{(\theta_2+\gamma)}+\hat{k}(\hat{D}-\hat{D}_0)\cos{(\theta_2+\beta)}\\
        +&\hat{b}\left( \cos{(\theta_1+\beta)}\theta_1' - \cos{(\theta_2+\beta)}\theta_2'\right) \cos{(\theta_2+\beta)}
    \end{split}
\end{equation}
\begin{equation}
\hat{D}' = \cos{(\theta_1+\beta)}\theta_1' - \cos{(\theta_2+\beta)}\theta_2'
\end{equation}
which have an augmented state vector
\begin{equation}
\hat{x} = \begin{bmatrix}
\theta_1&
\theta_1'&
\theta_2&
\theta_2'&
\hat{D}
\end{bmatrix}^T
\end{equation}
and whose continuous dynamics are only dependent on the parameters $\gamma$, $\hat{D}_0$ $\hat{k}$, and $\hat{b}$ and hybrid switching dependent on $\alpha$. The discrete jumps for the nondimensionalized hybrid dynamical system are identical to those in Section \ref{coupledModel}. 
\subsection{Acquisition of Optimal Coupling Parameters via Numerical Poincar\'e Return Map}
We now aim to find the stiffness $k$ and damping $b$ with the highest rate of convergence to synchronize a given system with a set of mass and length parameters. We will consider only the effects of $\hat{k}$ and $\hat{b}$ on the nondimensionalized system, which can then be used to find the physical stiffness $k$ and damping $b$ after scaling using physical parameters. 

Due to our system's behavior as a stable limit cycle oscillator, we will analyze the convergence of $\theta_2$ from one oscillation period to the next using a Poincar\'e return map. Consider a state-determined stable limit cycle oscillator with a state $x\in\Re^n$. There is a periodic orbit in the phase plane of this system. We define a Surface of Section (which can be an $n-1$ dimensional hyperplane transverse to the stable orbit at some specific point in the trajectory) through which intersect all trajectories converging to the stable limit cycle. If this hyperplane is orthonormal to one state variable in the phase plane, then the intersecting points are part of a reduced dimension Poincar\'e state vector $x_p\in\Re^{n-1}$. A stable fixed point of this discrete Poincar\'e system implies a stable limit cycle of the original system. 

A Poincar\'e return map $P()$ is an autonomous function representing the reduced state after one orbit to timestep $k+1$ given the state at time $k$
\begin{equation}
x_p[k+1]=P(x_p[k])
\end{equation}

The Poincar\'e return map $P()$ can be linearized about an equilibrium as
\begin{equation}
x_p[k+1]=\frac{dP(x_p[k])}{dx_p}x_p[k]=Ax_p[k]
\end{equation}
Because this is a discrete system, the eigenvalues $\lambda_i$ of matrix $A$ represent the dynamic response of the linearized system, which informs the local dynamic response of $P()$. If $|\lambda_i|<1, ~~i\in[1:n-1]$ then the discrete system is stable. The state $x_p$ at time $m$ given some initial condition $x_p[0]$ can be linearly approximated as
\begin{equation}
x_p[m] = A^m x_p[0]
\end{equation}

For the Nondimensionalized Dynamically Coupled Double Rimless Wheel system, we have a continuous time state vector
\begin{equation}
x= 
\begin{bmatrix}
\theta_1&
\theta_1'&
\theta_2&
\theta_2'&
\hat{D}
\end{bmatrix}^T
\end{equation}
We take a Surface of Section during the transition when $\theta_1=\alpha$ and takes a step. We wish for $\theta_2$ to converge to $0^\circ$ along this Surface of Section. Our augmented state vector is now
\begin{equation}
x_p = \begin{bmatrix}
\theta_1'&
\theta_2&
\theta_2'&
\hat{D}
\end{bmatrix}^T
\end{equation}

\begin{figure}[h]
	\centering
	\includegraphics[width=1\linewidth]{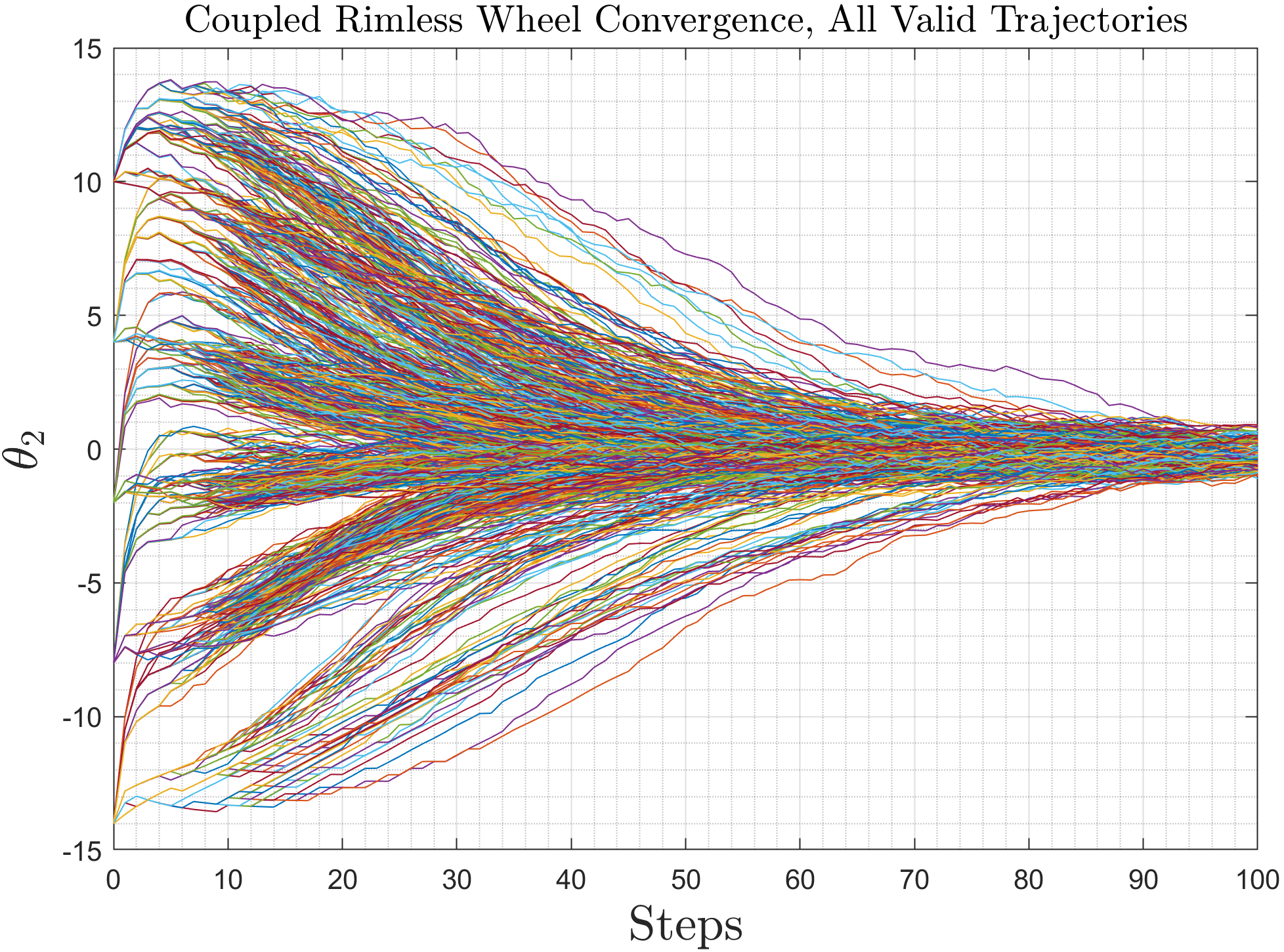}
	\caption{All converging trajectories from the set of numerical simulations.}
	\label{AllTrajFig}
\end{figure} 

The analytical solution to the time response of \eqref{eqSS1}, \eqref{eqSS2}, and \eqref{eqSS3} is dependent on knowing the time spent in both cases of continuous motion, which we do not know a priori. Thus, the analytical Poincar\'e return map for the Coupled Rimless Wheels system can be intractable to solve for, but the linear matrix $A$ may be obtained numerically using data collected from several simulated trajectories. 

Consider a trajectory of $x_p$ from an initial condition $x_{p}[0]$ to some convergent value $x_{p}[M]$ after $M$ discrete samples. Now we run $K$ different trials of the same system at different initial conditions, for a total of $M\times K$ samples of $x$. These data may be used to formulate an $A$ matrix using Least Squares Regression, as adapted from Chapter 3 of \cite{Goodwin:2009:AFP:1643720}. 

We wish to find matrix $A$ that minimizes the squared error between our predicted value of $x_{pred}[k+1]=Ax[k]$ and the actual value $x[k+1]$.
We may formulate this as an optimization:
\begin{equation}
J=\frac{1}{2}\sum_{k=1}^{K}\sum_{m=1}^{M-1}(Ax_p[m]-x_p[m+1])^T(Ax_p[m]-x_p[m+1])
\end{equation}
Taking the derivative and setting it equal to a zero matrix
\begin{equation}
\frac{dJ}{dA}=0=\sum_{k=1}^{K}\sum_{m=1}^{M-1}(Ax_p[m]-x_p[m+1])x_p^T[m]
\end{equation}
we can distribute and rearrange (with the sums written in shorthand for brevity)
\begin{equation}
A\sum\sum x_p[m]x_p^T[m]=\sum\sum x_p[m+1]x_p^T[m]
\end{equation}
Defining 
\begin{equation}
    P^{-1}=\sum_{k=1}^{K}\sum_{m=1}^{M-1} x_p[m]x_p^T[m]
\end{equation}
and 
\begin{equation}
   B=\sum_{k=1}^{K}\sum_{m=1}^{M-1} x_p[m+1]x_p^T[m] 
\end{equation}
we obtain the final matrix
\begin{equation}
A=BP
\end{equation}
The eigenvalues of $A$ provide the system rate of convergence. By performing this operation for different system parameters and comparing the eigenvalues, the parameters that lead to the fastest convergence rate may be selected. 

\begin{figure}[t]
	\centering
	\includegraphics[width=1\linewidth]{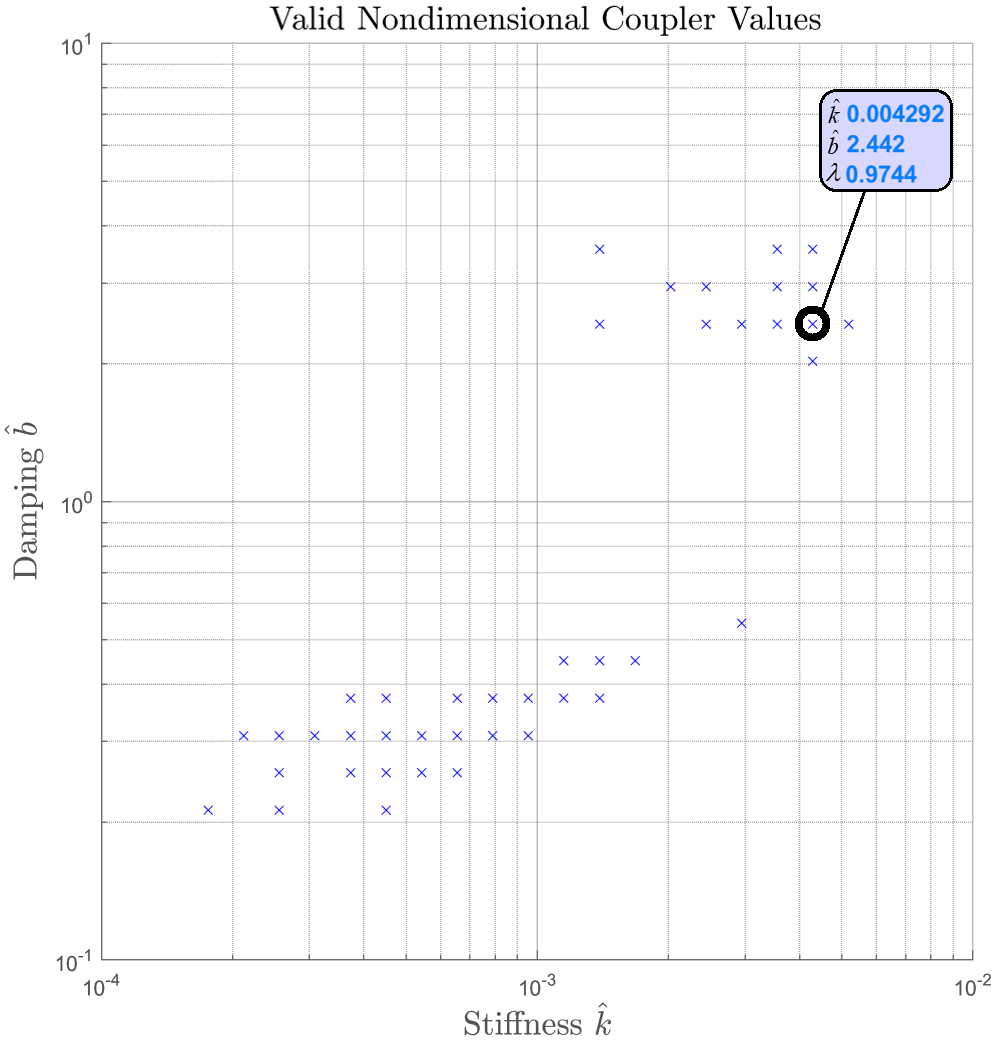}
	\caption{Valid combinations of $\hat{k}$ and $\hat{b}$ which lead to converging systems, and eigenvalue of the parameter set with fastest convergence. Note that because these are discrete systems, 1 is the threshold for stability, and the farther from 1 the eigenvalue is, the faster the response.}
	\label{PoincareEigs2Fig}
\end{figure} 

Data were obtained via physics simulations that were run for systems with a slope angle $\gamma=1.75^\circ$ nondimensional stiffness $\hat{k}$ from $10^{-4}$ to $10^{-2}$ and damping $\hat{b}$ from $10^{-1}$ to $10^1$. (Again, note that these parameters are nondimensional, but can be related to physical parameters using \eqref{paramNonDim}.) The same sets of initial conditions were used to generate the set of trajectories for each $[\hat{k},\hat{b}]$ pair. From these sets of trajectories, the valid $[\hat{k},\hat{b}]$ pairs were chosen. We consider valid combinations of $\hat{k}$ and $\hat{b}$ to be ones in which for all trajectories for all initial conditions $|\theta_2|<1^\circ$ within the simulation time. See Fig. \ref{AllTrajFig} and Fig. \ref{PoincareEigs2Fig}. Note that two separate regions of continuous points exist, which may be the result of our time cutoff. While there may be additional stable parameter sets that only converge given significantly more time, these points are by definition not operationally useful for our application.

An $A$ matrix was generated from each valid set of trajectory data corresponding to a valid $[\hat{k},\hat{b}]$ pair. The dominant (largest magnitude) eigenvalue of each $A$ was plotted for each $[\hat{k},\hat{b}]$ pair in Fig. \ref{PoincareEigs2Fig}.

\begin{figure}[ht]
	\centering
	\includegraphics[width=\linewidth]{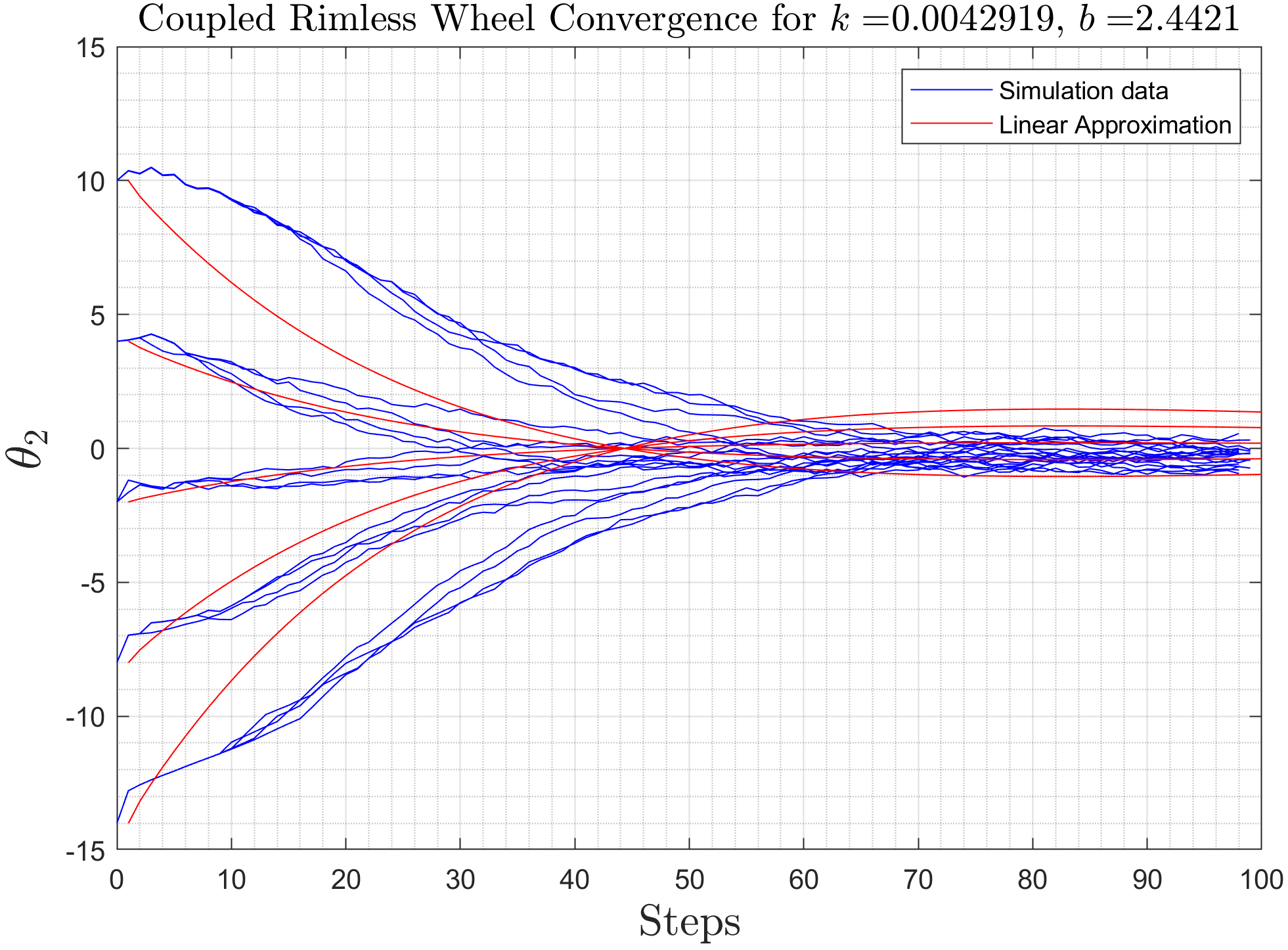}
	\caption{Converging trajectory for the best stiffness and damping selected from the set of numerical simulations.}
	\label{SingleTrajFig}
\end{figure} 

The $[\hat{k},\hat{b}]$ pair having the lowest magnitude discrete eigenvalue was selected. The simulation trajectories and the Least Squares estimate of the trajectories for this best parameter set are shown in Fig. \ref{SingleTrajFig}

\section{Experimental Validation of Passive Coupled Rimless Wheel Convergence to Gait Synchronization}\label{Experiment}
\begin{figure}[h!]
	\centering
	\includegraphics[width=0.875\linewidth]{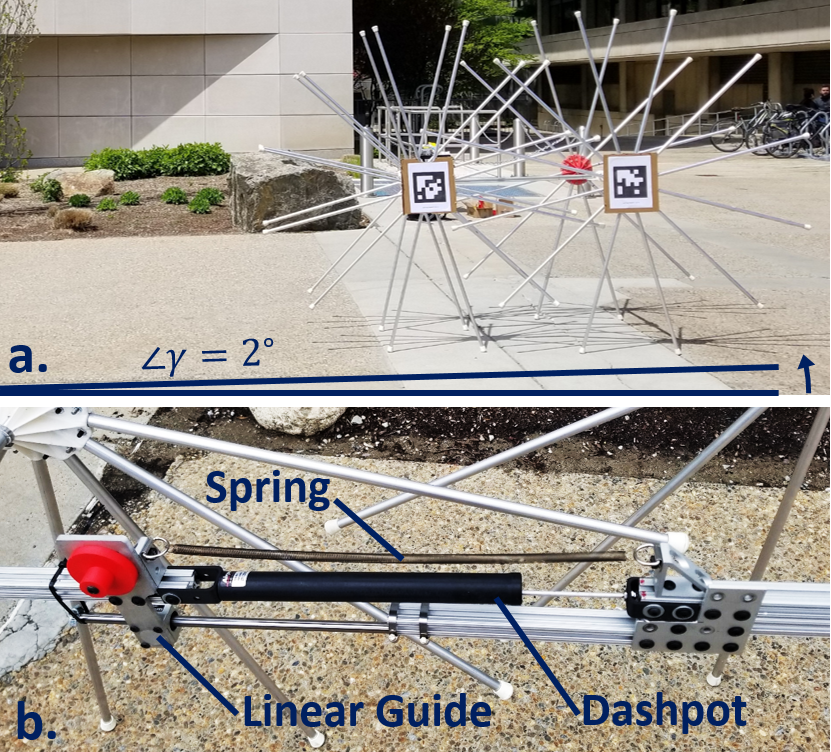}
	\caption{Setup for testing synchronization of the Coupled Rimless Wheels system using a hardware-implemented spring-dashpot coupler.}
	\label{figExperimentalSetup}
\end{figure}

A prototype Coupled Rimless Wheels system was built in order to test gait cycle convergence between two passive dynamic walkers (See Fig. \ref{figExperimentalSetup}). Each wheel was designed to be human-sized with $\ell=0.9652$ meters (38 inches) which is the center of mass for a 1.7272 meter (5 foot 8 inch) tall male. Twelve spokes give each wheel a step angle $\alpha = 15^\circ$.
\begin{figure*}[t!]
	\centering
	\includegraphics[width=0.875\textwidth]{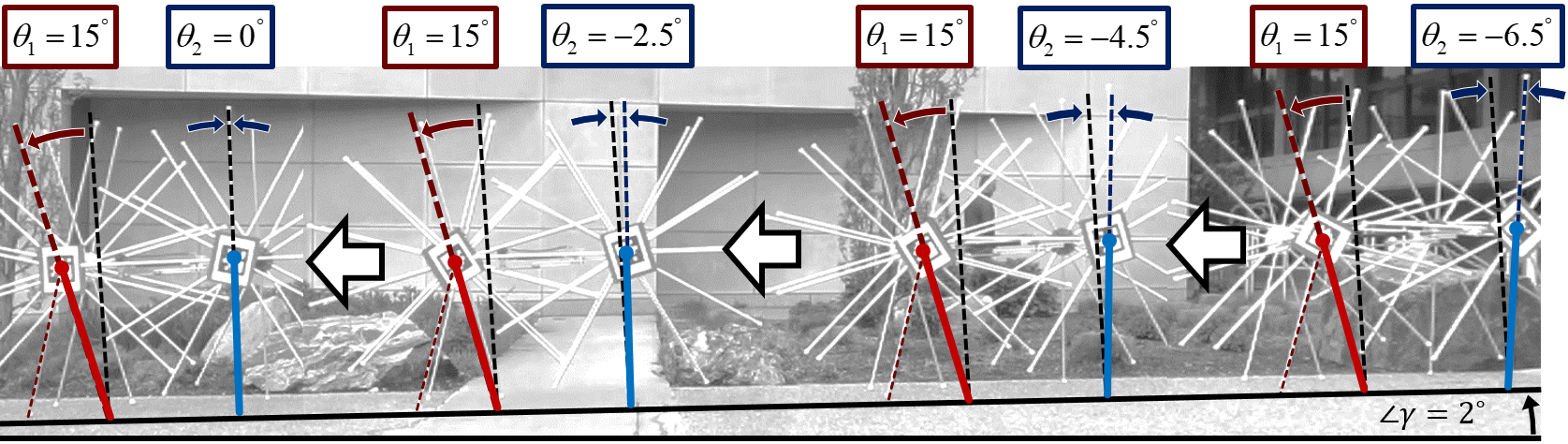}
	\caption{Synchronization of a physically implemented Coupled Rimless Wheels system. The system synchronizes to the desired gait cycle within 7 steps.}
	\label{figExperimentalTimeline}
\end{figure*}

Each Rimless Wheel consists of an identical pair of wheels, each with a hub made from ABS plastic, 12 spokes made of aluminum with rubber caps at the end of each for ground contact, and a rigid shaft, also aluminum, about which the coupler is free to rotate via a bearing. In order to track the position and orientation of each Rimless Wheel, AprilTag markers were mounted to the side of each wheel \cite{Olson2011}. 

The coupler consists of a spring and a dashpot constrained to be loaded only linearly. A custom dashpot was made by the Airpot Corporation with a 280mm (11 inch) stroke and a hand-adjustable damping range from 0 [Ns/m] to 5234 [Ns/m]. The damper was tuned to be roughly 100 [Ns/m]. In order to overcome static friction in the dashpot, and due to the lack of practical availability of springs of lower stiffness, the coupler spring was chosen to be 5.25 [N/m]. See Fig. \ref{figExperimentalSetup} for details.

The Coupled Rimless Wheels were sent down a gentle slope of $\gamma=2^\circ$ while a camera on a tripod filmed the result from the left side. Fig. \ref{figExperimentalTimeline} shows snapshots of the wheel's progress down the slope. The wheel phase converges to oscillate within $\pm 6\%$ of the desired phase difference of $50\%$ from an initial condition of $21.6\%$ within 7 steps. The difference in number of convergence steps between the simulation in Fig. \ref{SingleTrajFig} and the physical system can be attributed to a different slope angle $\gamma$ and unmodeled dynamics such as elasticity in ground collisions and friction in the coupler bearing. 

\begin{figure}[h!]
	\centering
	\includegraphics[width=0.9\linewidth]{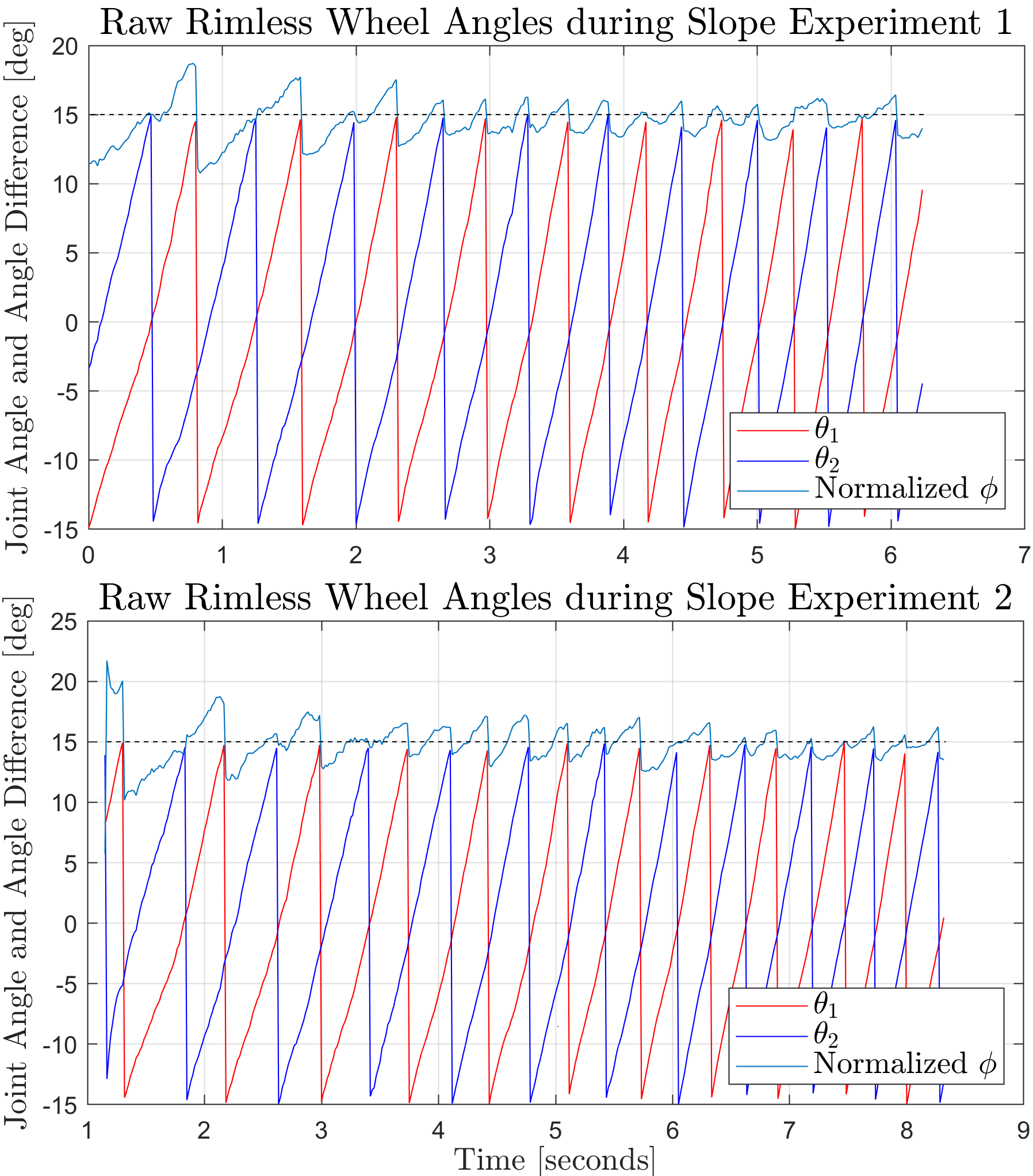}
	\caption{$\theta_1$, $\theta_2$, and $\phi$ during Experiment Trials 1 and 2}
	\label{figExperimentalResultsBoth}
\end{figure}

Fig. \ref{figExperimentalResultsBoth} shows the trajectory of the Rimless Wheels and the normalized angle difference
\begin{equation}
    \phi = \theta_1-\theta_2
\end{equation}
for the first 5 seconds of Trials 1 and 2. While $\phi$ is oscillatory for both trials, it converges to within $\pm2^\circ$ of the desired angle difference of $15^\circ$ within several steps. These results validate the convergence of two coupled walking systems through passive means alone.

\section{Conclusion and Recommendations for Future Work}\label{Conclusion}
This work explored the use of a passive spring-dashpot to couple two hybrid dynamic limit cycle oscillators and synchronize their motion, with an application in coordinating the gait of the human-XRL quadrupedal system during steady state locomotion. A novel Dynamically Coupled Double Rimless Wheel model was formulated to capture these dynamic interactions, and it was shown that the system converges to a desired gait cycle for certain coupler parameters. A Poincar\`e map was numerically acquired and used to identify the coupler parameters leading to the fastest synchronization convergence. A physical pair of Rimless Wheel walkers was built and coupled with a spring-dashpot matching these parameter values. Synchronization was observed when this physical system walked down a gentle slope.

This truly passive method using a dynamic coupler is rather limited in flexibility and adaptability. To extend the utility, active control will be considered. This active control will be built on the basis of the intrinsically contracting dual biped systems. Rather than fighting against the intrinsic dynamics, it will exploit the natural properties of the system. Additional analysis of necessary and sufficient conditions for synchronized steady-state walking will be performed, as well as thorough experimentation to further validate our findings.
\clearpage
\bibliographystyle{IEEEtran}
\bibliography{IEEEabrv,biblio}

\end{document}